\documentclass{article}

\usepackage{microtype}
\usepackage{graphicx}
\usepackage{subfigure}
\usepackage{booktabs} %
\usepackage{amsmath}

\usepackage{booktabs}
\usepackage{siunitx}

\usepackage{tikz}
\def\checkmark{\tikz\fill[scale=0.4](0,.35) -- (.25,0) -- (1,.7) -- (.25,.15) -- cycle;}

\usepackage{hyperref}

\usepackage[preprint]{mlsys2025}

\mlsystitlerunning{Wattribution: Every Joule, Every Layer}

\begin{document}

\twocolumn[
\mlsystitle{Wattribution: Every Joule, Every Layer --- Tracing and Predicting the Energy of Language Models}

\begin{mlsysauthorlist}
\mlsysauthor{Adrien Sardi}{nokia,inria,i3s}
\mlsysauthor{Marie-Line Alberi-Morel}{nokia}
\mlsysauthor{Sara Alouf}{inria}
\mlsysauthor{Frédéric Giroire}{i3s}
\mlsysauthor{Joanna Moulierac}{i3s}
\end{mlsysauthorlist}

\mlsysaffiliation{nokia}{Bell Labs, Nokia Networks France}
\mlsysaffiliation{inria}{Inria, Université Côte d'Azur, France}
\mlsysaffiliation{i3s}{Université Côte d'Azur, CNRS, Inria, I3S, France}

\mlsyscorrespondingauthor{Adrien Sardi}{adrien.sardi@nokia.com}

\mlsyskeywords{Energy Estimation, Layer Decomposition, Energy Measurement, Large Language Model, Inference, MLSys}

\vskip 0.3in

\begin{abstract}
Choosing or deploying a language model requires knowing its energy consumption in advance. Layer-wise estimators predict it from the architecture, but are trained on layers measured in isolation, under GPU conditions that differ from those of the complete model.
We introduce Wattribution, a protocol and tool that attributes the energy of a language model to each of its layers within its actual execution, building on the buffer of timestamped NVML power samples, which we characterize for the first time to our knowledge on nine GPUs from Turing to Blackwell, and which provides a new reading every 20~ms instead of a one-second average.
Using Wattribution, we collect nearly 34M per-layer values from 11 language models (0.6B to 8B parameters) over 1{,}548 model--configuration runs; these values are repeatable, add up to the energy of the full generation without any correction factor, and show that the prefill can account for two thirds of the energy of a run with long prompts.
A lightweight layer-wise estimator trained on these measurements predicts the energy of a language model from its configuration alone, with a median absolute percentage error of 4.8\% on unseen configurations, 13.1\% on language models never seen during training, and 16.9\% on unseen model families.
\end{abstract}
]

\printAffiliationsAndNotice{}  %

\section{Introduction}

The rapid adoption of Artificial Intelligence (AI) has made it a significant driver of electricity demand. Data centers already account for a notable share of global electricity consumption, and AI is expected to be the main driver of their growth in the coming years \cite{iea_energy_2025}. This growth is increasingly driven by the use of models rather than by their development, since a model is trained a single time but queried millions of times after deployment. Inference now represents a growing, and often dominant, share of the energy spent on AI.

Large language models (LLMs) are at the center of this shift. In a few years, they have moved from research prototypes to tools used daily by millions of people. This transformation is not limited to individual use. LLMs are increasingly embedded in automated pipelines, where agentic systems orchestrate many model calls to complete a single task, multiplying the number of inferences behind each user request.
AI is also expected to spread to every type of hardware: from local inference on personal devices, to base stations and other nodes of telecommunication networks, up to enterprise clusters. Each of these platforms has its own energy budget, and some of them, such as network nodes, operate under strict power constraints. In this context, knowing the energy consumption of an algorithm before deploying it becomes essential, whether to choose a model, size a platform, or decide where to place a workload in a network.

Yet there is no simple, standardized way to access this information. Accuracy and latency are routinely reported and used to compare models and make deployment decisions; energy consumption is not. 
Measuring it requires access to the target hardware, an instrumented setup and many repetitions. 
We argue that energy consumption must become a first-class metric to guide architectural and deployment decisions,
in the same way as accuracy and latency.

In this paper, we propose a reproducible and transferable approach to measure the energy consumption of the sub-modules that compose a neural network, and to use these measurements to estimate the energy consumption of a language model without running it.
To the best of our knowledge, this is the first method that measures the energy of each layer of a language model during its actual execution within the forward pass, and the first layer-wise energy estimator for language models built from such measurements.
Working at the layer level gives the estimator a finer granularity than architecture-level approaches, enabling transfer to unseen architectures while keeping it lightweight.
We structure this work around three research questions:
\newline

\textbf{RQ1} How can the energy consumption of a language model be measured accurately?

\textbf{RQ2} What are the main trends and mechanisms behind the energy consumption of a language model?

\textbf{RQ3} How can we predict, without running it, how much energy a given language model will consume?

Our key contributions are as follows:

1) \textbf{Power sensor characterization.} We study the power readings that NVIDIA exposes through NVML in the context of short GPU workloads, and measure their time resolution on nine GPUs from Turing to Blackwell. This study includes, to our knowledge, the first characterization of the sample buffer \texttt{nvmlDeviceGetSamples}, the reading with the highest resolution, and leads to practical guidelines for measuring the energy of sub-second processes.

2) \textbf{Per-layer measurement protocol.} Building on this sample buffer, we propose \textbf{Wattribution}, an experimental protocol and software tool to measure the energy consumption of neural network layers during their actual execution in the forward pass. 
We use it to collect energy data on 11 language models (0.6B to 8B parameters) across 200 inference configurations, totalling 1{,}548 model--configuration runs and nearly 34M per-layer energy measurements. 
To the best of our knowledge, this is the largest collection of per-layer energy data of language model inference, and the first to cover a controlled grid of inference parameters across models at this granularity.

3) \textbf{Energy characterization.} Leveraging the collected data, we analyze the energy consumption of language models and identify the main factors that drive it and that matter for its estimation.
The fine-grained data of Wattribution allow us to split the energy of each run,
first between the prefill and decode phases, and down to the granularity of individual layers.

4) \textbf{Layer-wise estimation model.} We build a lightweight, layer-wise energy estimation model, based on features computed
analytically from the model configuration. 
It predicts the energy consumption of a language model with a median absolute percentage error of 4.8\% on unseen configurations, 13.1\% on language models never seen during training, and 16.9\% on model families never seen during training. All this, evaluated with cross-configuration, leave-one-model-out and leave-one-family-out validation.

\section{Related Work}
\label{sec:related_work}

\subsection{Language Models}

Natural Language Processing (NLP) is a sub-field of machine learning that
develops models for language tasks, from text classification to text
generation. 
In 2014, encoder--decoder architectures were introduced for
sequence-to-sequence tasks such as machine translation
\cite{cho_etal_2014_properties}. Three years later, the Transformer
architecture \cite{vaswani_attention_2017} became the cornerstone of modern
language models, most of which now rely on its decoder-only variant. 
Like any neural network, a language model can be decomposed into blocks and layers, each performing a well-defined mathematical operation during the execution of the model.

\subsection{Measuring Energy}
\label{section:measurement}

The energy of a workload is obtained by sampling its power draw and integrating it over the execution. External power meters give the most reliable readings, but they require physical access to the hardware, are rarely available in shared or cloud infrastructures, and usually measure a whole device or node. On-board sensors, in contrast, are read through software interfaces and measure each component separately.
On NVIDIA GPUs, the power sensor is accessed through the NVIDIA Management Library (NVML) \cite{nvidia_nvml_2026}, on which \texttt{nvidia-smi} is built. Most energy measurement tools, such as CodeCarbon \cite{benoit_courty_2024}, Carbontracker \cite{anthony_carbontracker_2020}, Zeus \cite{you_zeus_2023} and Alumet \cite{raffin_hal_05246933}, rely on these interfaces, reading either the default power reading of NVML or its cumulative energy counter.

The quality of these readings has been studied since the introduction of the sensor. On early generations, the default reading was refreshed about every 20~ms, and \cite{lang_high_resolution_2013} exploited this rate to profile short GPU functions by repeating them at different phases of the update cycle. \cite{yang_accurate_2024} later characterized the sensor on more than 70 GPUs from Fermi to Hopper. 
They showed that, from Turing onward, the readings of \texttt{nvidia-smi} are refreshed only every 100~ms and, on A100 and H100, average over just 25~ms, leaving 75\% of the runtime unmeasured; they also estimated the error of the power reading at $\pm 5\%$, rather than the $\pm 5$~W stated by NVIDIA. \cite{dauner_evaluating_2026} found that the cumulative energy counter is no faster, updating at best every 100~ms, and can underestimate the energy by up to 95\% on consumer GPUs. Neither study evaluates the buffer of timestamped samples that NVML also exposes through \texttt{nvmlDeviceGetSamples}, which we characterize in Section~\ref{sec:sensor}.

\subsection{Measuring the Energy Consumption of AI}
Choosing or sizing a model requires knowing its energy consumption before deployment; model selection alone could reduce the worldwide energy consumption of AI by 27.8\% \cite{barros_small_2025}. 
Recent studies therefore measure the energy of AI models by executing them.
\cite{luccioni_power_2024} measure 88 models across 10 tasks and find generative tasks the most energy-intensive, while \cite{argerich_measuring_2024} show that the GPU accounts for more than 85\% of the energy of an LLM-based application. 
Other studies vary the inference setting.
\cite{zine_hal_05567430} analyze three models across 254 configurations, and show that energy consumption and latency are strongly correlated and mainly driven by options that increase the number of forward passes, such as offloading the KV cache. 
Conversely, \cite{reichwald_mapping_2026} benchmark 75 small language models in a single configuration with a batch size of one. They study the trade-offs between model size, task performance and energy consumption, and how these have evolved from 2023 to 2025.
These studies either explore many configurations of a few models, or many models in a single configuration, and their findings remain tied to the measured models. Our data collection, which targets models of a similar size, lies between these two extremes, varying both the models and the experimental parameters, so that the energy of unseen architectures can be estimated without execution.

\subsection{Estimating the Energy Consumption of AI}

The simplest estimates of the energy consumption of a model rely on proxies such as its number of parameters or of FLOPs. These proxies are useful to compare models within the same family, but they are not accurate estimators: they ignore how a model is executed, as well as the data movements between and within hardware components \cite{sze_efficient_2017}. 
Data-related metrics, such as the number of activations, have shown better results \cite{bytezcom_icml_2025}. Another line of work estimates the energy of a neural network from the energy of its sub-processes, such as blocks or layers \cite{cai_neuralpower_2017, cao_etal_2021_irene, yao_eali_2022, getzner_accuracy_2023, zhang_ampere_2025, sardi2026wattlayer}. These approaches have mostly been applied to classification networks, mainly in computer vision, and, to our knowledge, none of them addresses the autoregressive generation of language models.

Building such an estimator requires measuring the energy of each sub-process. Existing approaches execute layers in isolation, which raises two issues. First, a layer often runs for a shorter time than the update period of the power sensor (Section~\ref{section:measurement}), which leads to measurement errors that propagate to the estimates \cite{yao_eali_2022}. Second, measuring a layer in isolation assumes that
it consumes the same energy as within the complete model. This assumption does not hold on GPUs, whose clock frequency depends on the surrounding workload: a layer executed alone runs under different conditions and typically consumes less energy than within the full architecture (Appendix~\ref{app:aggregation-gap}). As a result, the sum of the layer energies underestimates the energy of the model and must be corrected empirically \cite{cao_etal_2021_irene, sardi2026wattlayer}. EALI \cite{yao_eali_2022} addresses this issue by fixing the GPU frequency and using it as a feature of the estimator. However, apart from settings where the provider locks the clocks, the frequency at which a GPU will execute a model can only be known by running it, which prevents its use for estimation before execution. 
Wattribution avoids both issues by measuring each layer within the actual execution of the complete model, which also splits the energy of each run between the prefill and decode phases.

\section{Methodology}
\label{sec:methodology}

\subsection{Experimental Setting}
\label{sec:setting}

We focus on the energy consumed by the GPU, which accounts for most of the energy of LLM inference, up to 85\% \cite{argerich_measuring_2024}. 
To measure it, we rely on the NVIDIA Management Library (NVML), the library on which \texttt{nvidia-smi} is built. NVML reads the GPU power from an on-board power sensor \cite{yang_accurate_2024} and exposes it through five readings, which differ in how often they are updated and over which interval they average. 
Most of the measurement tools rely either on the default power reading, \texttt{nvmlDeviceGetPowerUsage}, or on the cumulative energy counter \cite{benoit_courty_2024,raffin_hal_05246933,you_zeus_2023}. On recent GPUs, the former averages the power over about one second \cite{yang_accurate_2024}, and the latter can strongly underestimate the energy on some devices \cite{dauner_evaluating_2026}. Wattribution instead reads the buffer of timestamped power samples returned by \texttt{nvmlDeviceGetSamples}, which provides a new value about every 20~ms. We justify this choice by characterizing the five readings on nine GPUs from the Turing, Ampere, Hopper, Ada Lovelace and Blackwell generations (Section~\ref{sec:sensor}). Querying NVML directly gives us full control over the measurement process and makes our tooling compatible with any NVIDIA GPU, which eases future extensions of this work.

The data collection campaign is run on a workstation equipped with an NVIDIA RTX 6000 Ada Generation GPU.
\begin{table}[t]
  \centering
  \small
  \caption{Configuration space and profiled models. \#Cfg is the number of configurations
  measured per model (out of 200); missing combinations exceeded GPU memory.}
  \label{tab:setup}
  \small
  \begin{tabular*}{\columnwidth}{@{\extracolsep{\fill}}llr}
    \toprule
    \multicolumn{3}{@{}l}{\textbf{Configuration parameters}} \\
    \midrule
    Input length      & \multicolumn{2}{l}{16 -- 8{,}192 tokens (8 values)} \\
    Generated tokens  & \multicolumn{2}{l}{16, 64, 128} \\
    Batch size        & \multicolumn{2}{l}{1, 32, 64, 128} \\
    Precision         & \multicolumn{2}{l}{fp32, fp16, bf16} \\
    \midrule
    \textbf{Family} & \textbf{Model} & \textbf{\#Cfg} \\
    \midrule
    Qwen3      & Qwen3-0.6B              & 200 \\
               & Qwen3-1.7B              & 178 \\
               & Qwen3-4B                & 156 \\
               & Qwen3-8B                & 144 \\
    \addlinespace
    Llama      & Llama-3.2-1B-Instruct   & 191 \\
               & Llama-3.2-3B-Instruct   & 158 \\
               & Llama-3.1-8B-Instruct   & 137 \\
    \addlinespace
    Gemma 3    & gemma-3-1b-it           & 107 \\
               & gemma-3-4b-it           &  95 \\
    \addlinespace
    Mistral    & Mistral-7B-Instruct-v0.3 &  84 \\
    Phi        & Phi-4-mini-instruct     &  98 \\
    \midrule
    \multicolumn{2}{@{}l}{Total runs} & 1{,}548 \\
    \bottomrule
  \end{tabular*}
\end{table}
Each experiment is defined by a configuration of four parameters: input length, number of generated tokens, batch size and numerical precision (Table~\ref{tab:setup}). We profile eleven decoder-only LLMs from five families, ranging from 0.6B to 8B parameters. Of the 288 possible configurations, 200 were measured, for a total of 1,548 runs; the remaining combinations exceeded GPU memory, mostly at large batch sizes and long inputs on the larger models, which explains the varying coverage per model.
Their selection balances the two needs of the estimation model, between studies of many configurations of a few models and of many models in a single configuration (Section~\ref{sec:related_work}): training it requires variety in the experimental parameters, and evaluating its transferability requires a sufficiently large number of models.

\subsection{Wattribution protocol}

\label{sec:wattribution}

This section describes Wattribution, the protocol we use to measure the energy consumption of the layers of a neural network. Wattribution is also the name of the software tool that implements this protocol.
The objective is to measure the energy of a sub-process, here a layer, inside the execution of a complete model. This is not straightforward.
Like any energy measurement, the energy of a layer is the product of its duration and of the power drawn while it runs: $E_{layer} = T_{layer} \times P$. 
But a layer executes in less than a millisecond, faster than the GPU power sensor refreshes. Two choices are then possible.
Measuring a layer in isolation, repeated many times, makes it the only work seen by the sensor, but it changes the operating point of the GPU: its clock frequency and voltage are set by the sustained workload, so an isolated layer runs at a frequency and power that do not occur during inference (Appendix~\ref{app:aggregation-gap}). The sum of isolated layer energies then misses the energy of the model, and must be corrected statistically \cite{cao_etal_2021_irene, sardi2026wattlayer}.
Measuring the layer within the complete model keeps the operating point that the GPU sets for the actual workload, which consecutive layers share.
Wattribution makes this second choice: every layer is measured in the actual inference, with the clock, voltage and memory state that the GPU sets for the complete model. 
Wattribution records the 
time window of each layer as seen by the host
and resolves the power at the scale of the sensor window, so that the energies of the layers add up to the energy of the model without any correction. A rigorous definition of what we measure and how we attribute it is therefore required.

\begin{figure}[!b]
    \centering
    \includegraphics[width=.9\columnwidth]{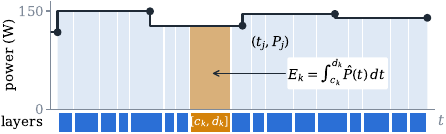}
    \caption{Attribution in Wattribution. Each power reading $(t_j, P_j)$ is held
    constant over the interval that precedes its timestamp (backward hold).
    The energy of a firing is the integral of this reconstructed power over
    its attribution window $[c_k, d_k]$ (highlighted). Window widths and
    power values are illustrative.}
    \label{fig:wattribution_rules}
\end{figure}

\textbf{Layers, firings and attribution windows.}
Let $\mathcal{A}$ be an architecture with $N$ layers $L_1, \dots, L_N$. Throughout this work, a \emph{layer} is a module of the model without children, such as a linear projection, a normalization or an activation function, and not a whole Transformer block.
We measure the generation of $T$ new tokens, and call each complete execution of this generation a \emph{repetition}. Within a repetition, each layer executes once during prefill (step $s = 0$) and once at each decode step ($s = 1, \dots, T-1$). We call each execution of a layer a \emph{firing}. A repetition therefore contains $K = NT$ firings, indexed $k = 1, \dots, K$ in execution order. For firing $k$, 
forward hooks record its start time $c_k$ and end time $d_k$ on the host monotonic clock.
In parallel, a background thread collects the GPU power trace $\{(t_j, P_j)\}_{j=1}^{M}$ through \texttt{nvmlDeviceGetSamples}, timestamped on the same host clock as the hooks. 
Each reading $P_j$ is an average of the power over a short window, and a new reading is produced about every 20~ms. 

\textbf{Attribution rules.}
An attribution rule reconstructs the power inside each window from the discrete trace, and the energy of firing $k$ is the integral of the
reconstructed power $\hat{P}$ over its window. 
We reconstruct the power with a backward hold. Each reading is held constant over the interval
that precedes its timestamp (see Figure~\ref{fig:wattribution_rules}).
This choice follows the behavior of the \texttt{samples} reading, whose timestamp falls after the window it averages %
: a
reading therefore describes the power before its timestamp rather than after it.
\begin{equation}
E_k = \int_{c_k}^{d_k} \hat{P}(t)\,dt,
\quad
\hat{P}(t) = P_j \;\; \text{for } t \in (t_{j-1}, t_j].
\label{eq:attribution}
\end{equation}

\textbf{Aggregation.}
We write $E^{(r)}_{\ell,s}$ for the energy $E_k$ of the firing of layer $\ell$ at step $s$ of repetition $r$. The energy of the layer in each phase is averaged over the $R$ repetitions:
\vspace{-.5em}
\begin{equation}
\begin{aligned}
\bar{E}_{\ell,\text{prefill}} = \frac{1}{R} \sum_{r=1}^{R} E^{(r)}_{\ell,0}, \hspace{0.5em}
\bar{E}_{\ell,\text{decode}} = \frac{1}{R} \sum_{r=1}^{R} \sum_{s=1}^{T-1} E^{(r)}_{\ell,s}.
\end{aligned}
\label{eq:aggregation}
\end{equation}
A firing is shorter than the interval between two readings, so its energy depends on where it falls relative to them. 
This position changes from one repetition to the next, and averaging over repetitions removes this dependence.
Section~\ref{sec:eval_wattribution} verifies that this sum matches the measured energy of the forward pass.
\vspace{-.5em}
\begin{equation}
E_{\mathcal{A}}
= \sum_{\ell=1}^{N} \left( \bar{E}_{\ell,\text{prefill}} + \bar{E}_{\ell,\text{decode}} \right)
= E_{\mathcal{A}}^{\text{prefill}} + E_{\mathcal{A}}^{\text{decode}}.
\label{eq:architecture}
\end{equation}

\vspace{-.5em}

\subsection{Estimation Model}
\label{sec:estimation}

Wattribution provides the energy consumed by each layer of a model. To estimate the energy of a model without running it, we describe each layer by a set of features that can be known in advance, and we learn a function that maps these features to the measured energy.

\textbf{Features.}
We use two families of features. \emph{Experiment features} describe the inference setting: input length, number of generated tokens, batch size and
numerical precision. \emph{Model features} describe the work performed by each layer, both in computation (MACs) and in memory traffic (bytes of
weights, inputs and outputs, and their sum). They are computed analytically from the configuration of the model, without downloading its weights or
running it (Appendix~\ref{app:features}). Since prefill is mostly limited by computation and decode by memory traffic (Appendix~\ref{app:features_block}),
we compute every model feature separately for each phase, summing the decode features over the decode steps, as Wattribution does for the energy.
Section~\ref{sec:energy} confirms this distinction on the measured data.
 
\textbf{From layers to architecture.}
Let $\mathbf{x}_{\ell,\text{prefill}}$ and $\mathbf{x}_{\ell,\text{decode}}$ be the feature vectors of layer $\ell$ in each phase, and let $f$ be the learned estimator. Since Wattribution measures each layer within the execution of the complete model, the energy of the architecture is the sum of the energies of its layers (Equation~\ref{eq:architecture}). The estimate inherits this property:
No multiplicative coefficient is needed to move from the layers to the architecture, unlike the correction factor required when layers are measured in isolation \cite{cao_etal_2021_irene,sardi2026wattlayer}.
\vspace{-.5em}
\begin{equation}
\hat{E}_{\mathcal{A}} = \sum_{\ell=1}^{N} \left( f(\mathbf{x}_{\ell,\text{prefill}}) + f(\mathbf{x}_{\ell,\text{decode}}) \right).
\label{eq:estimate}
\end{equation}

\vspace{-.5em}
\textbf{Choice of the estimator.}
We compare several families of regressors: linear and log-linear regression, multilayer perceptrons, and tree-based methods (random forest, histogram-based gradient boosting and XGBoost). 
Our selection is guided by two properties. The estimator must be \emph{lightweight}, so that it can be evaluated instantly at design time. 
It must be \emph{transferable}, so that it predicts the energy of architectures that were never measured. 
The models are evaluated with the cross-validation scheme described in Section~\ref{sec:prediction}, including a leave-one-model-out and leave-one-family-out split.

\section{Evaluation}
\label{sec:evaluation}
 
We evaluate each component of our methodology and answer the three research questions. Section~\ref{sec:sensor} characterizes the power readings of NVML and motivates the use of the sample buffer. Section~\ref{sec:eval_wattribution} validates the Wattribution protocol. Together, they answer RQ1. Section~\ref{sec:energy} analyzes the energy consumption of language models and the factors that drive it (RQ2). Section~\ref{sec:prediction} evaluates the estimation model (RQ3).

\subsection{Power Sensor Characterization}
\label{sec:sensor}

Every per-layer energy figure inherits the properties of the power reading it is computed from. NVML exposes five readings: \texttt{usage} (\texttt{nvmlDeviceGetPowerUsage}, the default of most tools), \texttt{instant}, \texttt{average}, \texttt{samples} (\texttt{nvmlDeviceGetSamples}, a buffer of timestamped readings) and a cumulative \texttt{energy} counter. Prior studies have characterized the readings of \texttt{nvidia-smi} and the energy counter, but not the sample buffer~\cite{lang_high_resolution_2013, yang_accurate_2024,dauner_evaluating_2026}. 
Our campaign extends these studies in three directions. 
(i) It uses current drivers and covers the Blackwell generation. 
(ii) It includes one reading outside the scope of their study, the \texttt{samples} buffer. 
(iii) Finally, it evaluates the sensor during real language-model inference, on the workload measured by Wattribution, and not only under synthetic load.

\begin{figure}[t]
\centering
\begin{tikzpicture}[x=0.15cm, y=1cm, font=\scriptsize]
  \draw[->] (0,0) -- (52,0) node[right] {$t$};
  \fill[gray!35] (14,0) rectangle (39,0.35);
  \draw (14,0) rectangle (39,0.35);
  \node at (26.5,0.17) {power averaged};
  \draw[|<->|] (14,0.6) -- node[above] {window $W$} (39,0.6);
  \draw[|<->|] (39,0.6) -- node[above] {lag $L$} (48,0.6);
  \fill (48,0) circle (2pt);
  \fill (6,0) circle (2pt);
  \node[below, align=center] at (48,-0.05) {reading $k$\\at $t_k$};
  \node[below, align=center] at (6,-0.05) {reading $k{-}1$\\at $t_{k-1}$};
  \draw[|<->|] (6,-0.95) -- node[below] {update period $P$} (48,-0.95);
\end{tikzpicture}
\caption{Model of one sensor reading. The value stamped at $t_k$ is the mean
power over $[t_k - L - W,\; t_k - L]$. A new reading appears every $P$. For
\texttt{samples} on our GPU, $P = 19.9$\,ms, $W = 17$\,ms and $L = 6$\,ms.}
\label{fig:model}
\end{figure}
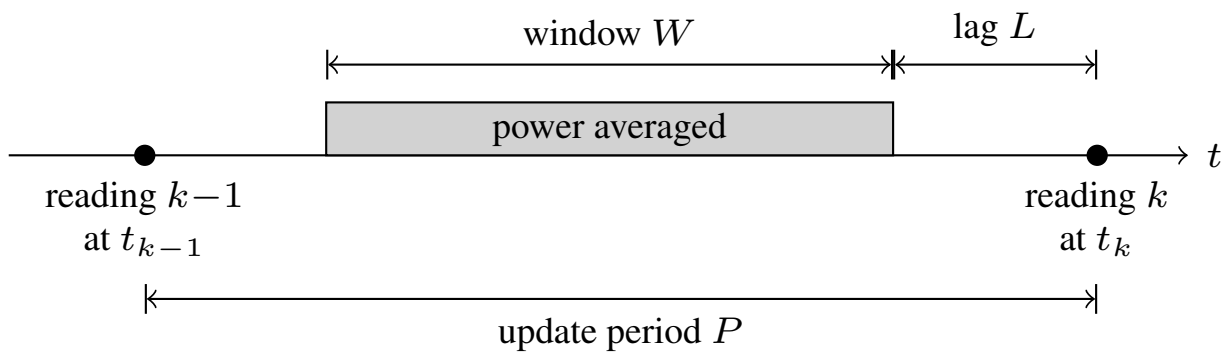

\textbf{What does a reading represent.}
A reading is not a snapshot of the power at an instant. It is an average over a recent stretch of time, described by three quantities (see Figure~\ref{fig:model}): the update period $P$, the time between two consecutive readings; the averaging window $W$, the length of the stretch it averages over; and the lag $L$, the gap between the end of that stretch and its timestamp. 
With $p(t)$ the true power, we model the reading stamped at $t_k$ as the mean power over a window of length $W$ that ends $L$ before its timestamp:
\begin{equation}
\hat{p}_k = \operatorname{mean}\bigl\{\, p(t) \;:\; t \in [t_k - L - W,\ t_k - L] \,\bigr\}.
\label{eq:reading}
\end{equation}
This is a phenomenological model: NVML does not document how the driver computes a reading, and we only use the model to estimate $W$ and $L$ from the observed readings.
The ratio $c = W/P$, which we call coverage, is the share of time actually measured. Again, this ratio is not documented by NVIDIA.

\textbf{Campaign.}
We characterize the five readings on nine GPUs from five generations, Turing, Ampere, Hopper, Ada Lovelace and Blackwell, with one unit per model (Table~\ref{tab:sensor}). 
Eight of them run driver 580.126.09 on a shared research cluster \cite{grid5000} and the reference workstation 
runs driver 595.84. 
A probe script polls the five readings in a loop at a fixed interval (2, 5 or 10~ms) while the GPU is in one of three known states: idle, under a constant fp16 matrix-multiplication load, or under a square-wave load that switches the same multiplication on and off with a 50\% duty cycle, for periods ranging from 10~ms to 4~s. 
Every transition is timestamped on the same host clock as the readings, so the true load signal is known. We estimate $W$ and $L$ by fitting Equation~\ref{eq:reading} to the square-wave response, with $p(t)$ modeled as a two-level signal, and keep the pair with the smallest residual (Appendix~\ref{app:sensor}).

\textbf{The sample buffer restores a 20~ms resolution.}
\texttt{samples} is the fastest reading on every GPU we tested (Table~\ref{tab:sensor}). It provides a new value about every 20~ms, averaged over 12 to 20~ms, with a coverage of at least 0.84 from Ampere onward. It thus brings back, on current GPUs, the 20~ms resolution that the default reading offered on earlier generations \cite{lang_high_resolution_2013, yang_accurate_2024}.
The other readings are much coarser. \texttt{usage} and \texttt{average} return the same value, an average over about one second refreshed every 100 to 500~ms. This explains why \cite{dauner_evaluating_2026} find the NVML power reading insensitive to the sampling interval: polling it faster returns the same smoothed value more often.
\texttt{instant} averages over about 100~ms at the same pace, so that up to 80\% of the runtime is never measured. The difference is visible under a square wave with a 100~ms period: \texttt{samples} follows each burst, whereas \texttt{usage} stays flat at the mean of the two levels (Figure~\ref{fig:square_wave}, Appendix \ref{app:sensor_readings}).
Finally, the \texttt{energy} counter brings no more information than \texttt{usage}. Where it is accurate, it agrees with \texttt{usage} and is updated only every 100~ms. On two of the nine GPUs, including the
reference workstation, it even reads far below \texttt{usage}, as \cite{dauner_evaluating_2026} observe on consumer GPUs. 
Although it is the default source of several measurement tools \cite{benoit_courty_2024,raffin_hal_05246933,you_zeus_2023}, we do not
use it.

\begin{table}[t]
\centering
\footnotesize
\caption{Power readings of NVML on nine GPUs, at 10~ms polling. $P_{\texttt{usage}}$ and $P_{\texttt{samples}}$ are the median update periods (in ms); $W$, $L$, $c$ and $R^2$ come from the square-wave fit on \texttt{samples}. All values in ms except $c$ and $R^2$. $^\dagger$Reference workstation, driver 595.84; the other GPUs run driver 580.126.09.}
\label{tab:sensor}
\setlength{\tabcolsep}{2.2pt}
\small
\begin{tabular}{@{}llrrrrrr@{}}
\toprule
\textbf{GPU} & \textbf{Arch.} & $P_{\texttt{usage}}$ & $P_{\texttt{samples}}$ & $W$ & $L$ & $c$ & $R^2$ \\
\midrule
Tesla T4             & Turing    & 102 & 20.3 & 12 & 4 & 0.59 & 0.70 \\
Quadro RTX 6000      & Turing    & 100 & 20.2 & 17 & 4 & 0.85 & 0.86 \\
A100 PCIe            & Ampere    & 100 & 20.2 & 17 & 1 & 0.84 & 0.89 \\
A40                  & Ampere    & 100 & 20.0 & 20 & 2 & 1.00 & 0.95 \\
H100 NVL             & Hopper    & 100 & 20.0 & 17 & 4 & 0.85 & 0.94 \\
L40                  & Ada       & 100 & 19.9 & 17 & 4 & 0.85 & 0.85 \\
L40S                 & Ada       & 100 & 20.0 & 20 & 4 & 1.00 & 0.93 \\
RTX 6000 Ada$^\dagger$ & Ada     & 500 & 19.9 & 17 & 6 & 0.85 & 0.87 \\
RTX PRO 6000         & Blackwell & 100 & 20.0 & 20 & 1 & 1.00 & 0.95 \\
\bottomrule
\end{tabular}
\end{table}

\textbf{Reading the sample buffer in practice.} We poll the buffer every 10~ms. 
Following the Nyquist criterion, polling twice as fast as the sensor updates is enough to capture every reading
\cite{bytezcom_icml_2025}. Polling faster only adds overhead: at 2~ms, the measured idle power rises by up to $3\times$.
The buffer also makes the measurement robust. 
When the polling thread is paused by the system, the readings produced in the meantime stay in the buffer and are retrieved at the next call. Only 0.2\% of about 115,000 readings are skipped or repeated, whereas with \texttt{usage}, every pause loses readings.

\subsection{Protocol Validation}
\label{sec:eval_wattribution}

We validate Wattribution against three properties of a trustworthy measurement, as no reference instrument measures the energy of a single layer within a model: the instrumentation overhead remains limited (P1), repeated measurements of a layer agree (P2), and the layer energies add up to the energy of the generation (P3). Together, they show that Wattribution describes the model rather than the instrument.

All experiments run on the reference workstation, with the \texttt{samples} reading polled every 10~ms, on seven models ranging from a small test model to Qwen3-8B. Each model generates 32 tokens from a 64-token prompt, with batch size 32 and fp16 precision. Each configuration is measured in 10 independent trials, executed in randomized order, each averaging 18 repetitions. Energies are reported net of the idle power of the GPU, measured before the model is loaded (32.5~W).
\textbf{P1: Limited overhead.} We quantify the overhead of Wattribution by interleaving runs with and without its hooks, both while sampling the power. 
The overhead amounts to a median of 6.9\% of the energy, small compared with the differences between layers on which the estimator is trained.
\textbf{P2: Repeatable layer energies.} %
Across trials, the median coefficient of variation of the layer energies is 2.3\%. This repeatability relies on averaging: since the sensor updates on its own clock, the position of a layer relative to the readings varies from one repetition to the next, which we verify with a Kolmogorov--Smirnov test against a uniform distribution for all 1\,593 layers, following the phase-shifting principle of \cite{lang_high_resolution_2013}. 
\textbf{P3: Additive layer energies.} The sum of the layer energies matches the measured energy of the generation within 0.7\%. This is the central difference with approaches that measure layers in isolation, where the sum does not match the energy of the complete model (see Appendix~\ref{app:aggregation-gap}).
By measuring each layer within the complete execution, Wattribution makes the energy of an architecture the sum of the energies of its layers.

\vspace{4pt}
\noindent\colorbox{gray!10}{\parbox{\dimexpr\columnwidth-2\fboxsep\relax}{%
\textbf{Answering RQ1: How can the energy consumption of a language model be measured accurately?}
Accurate measurement starts with the right power reading. 
The buffer of timestamped samples of NVML, which we characterize here for the first time to our knowledge, provides a new value about every 20~ms and covers most of the runtime on current GPUs.
Building on this reading, Wattribution attributes the measured energy to each layer during the actual execution of the model. The resulting layer energies are repeatable and additive, without any correction factor and at a limited instrumentation overhead. 
To our knowledge, Wattribution is the first protocol to attribute the energy of a language model to each of its layers within its actual execution, paving the way for finer-grained energy studies of neural networks.
}}

\subsection{What Drives the Energy of a Language Model?}
\label{sec:energy}

\setcounter{figure}{2}
\begin{figure}[!t]
    \centering
    \includegraphics[width=\linewidth
    , trim=0 22 0 0, clip
    ]{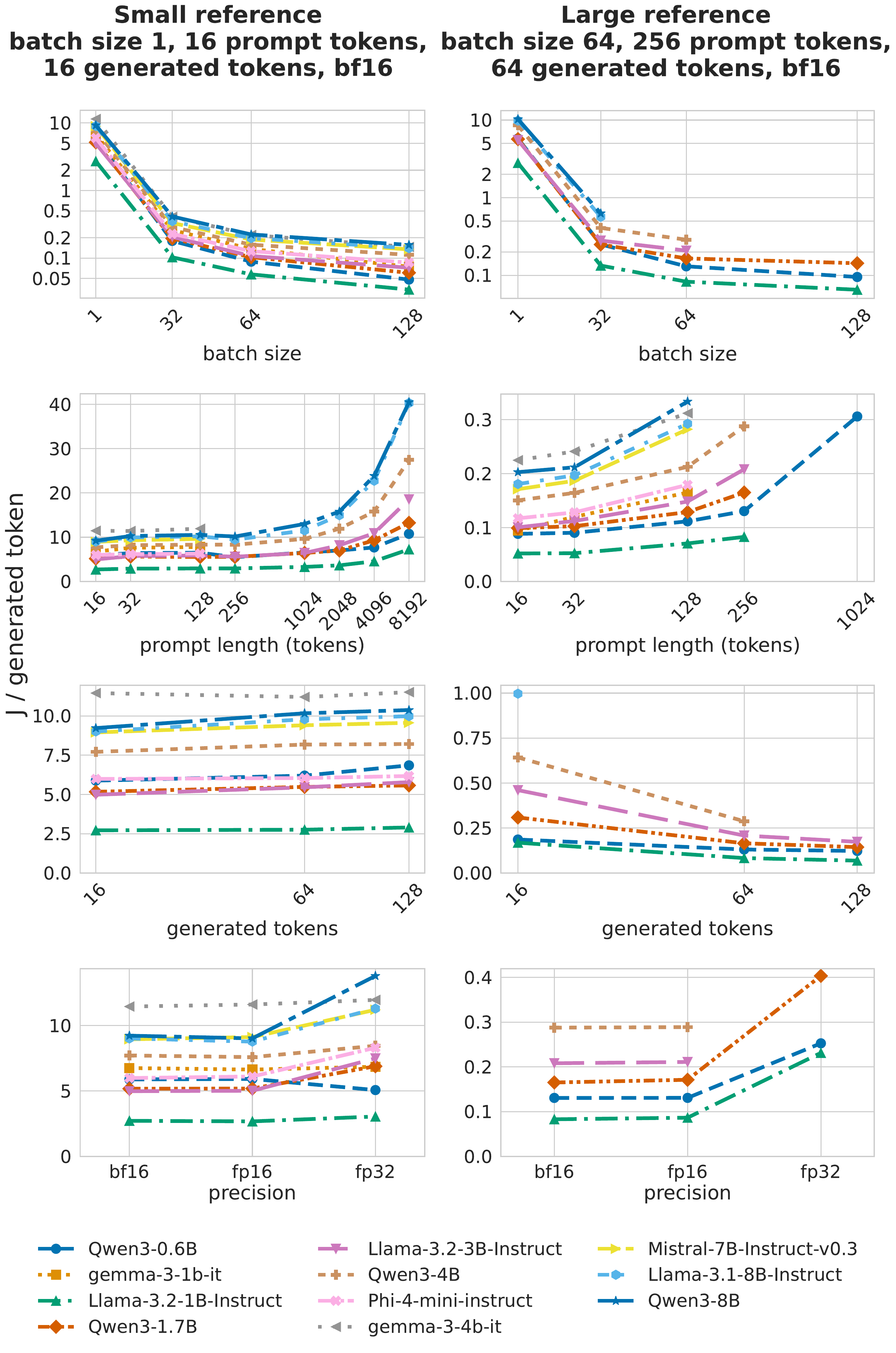}
    \caption{
    Energy per generated token when one factor varies and the other three stay at a reference configuration. 
    Missing points exceed the GPU memory. 
    }
    \label{fig:oat}
\end{figure}
\setcounter{figure}{3}
\begin{figure*}[t]
  \centering
  \includegraphics[width=.85\linewidth
  , trim=0 8 0 0, clip
    ]{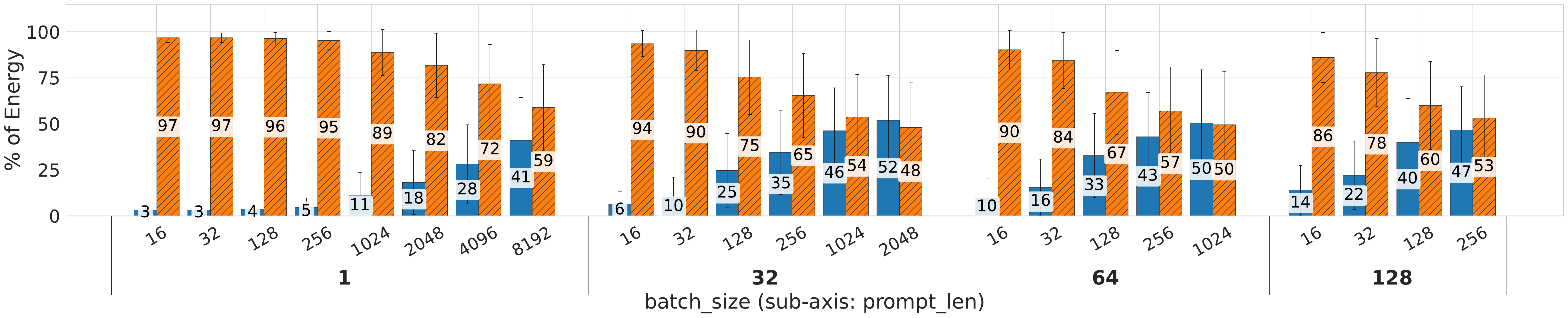}
  \par\medskip
  \includegraphics[width=.85\linewidth
  , trim=0 8 0 6, clip
    ]{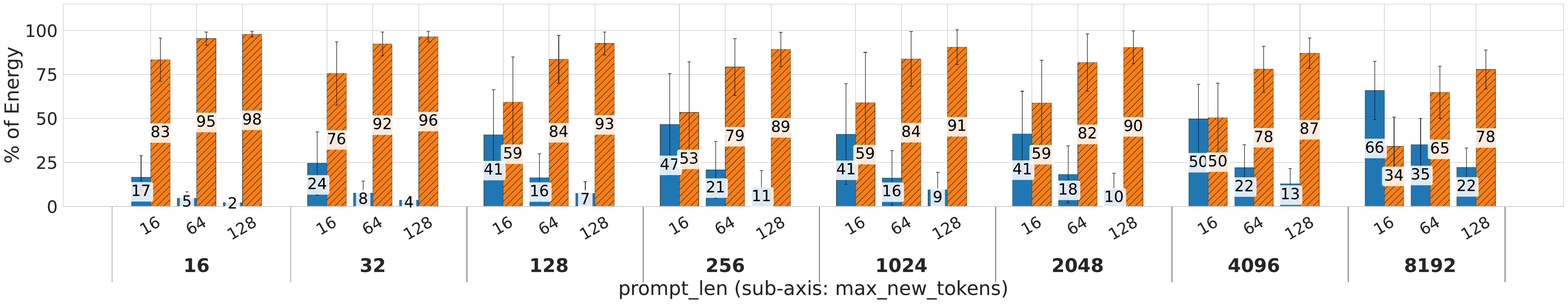}
  \caption{Share of the run energy spent in prefill (blue) and decode
  (orange). \textbf{Top:} by batch size and prompt length, all numbers of generated
  tokens pooled. \textbf{Bottom:} by prompt length and number of generated tokens,
  all batch sizes pooled. Each bar averages all the runs that share its two
  settings, across models and the remaining settings.}
  \label{fig:phase_share}
\end{figure*}

With the protocol validated
We now use IT to answer RQ2:
\emph{What are the main trends and mechanisms behind the energy consumption
of a language model?}
We focus on one metric, the energy per generated token (J/token). Tokens are
the unit that every language model takes as input and produces as output,
which gives a common basis to compare models with each other. It is also the
unit in which inference is priced by serving providers, often with separate
rates for input and output tokens.
We study 11 language models on the RTX~6000 Ada at two levels. 
At the architecture level, we measure how batch size, prompt
length, number of generated tokens and precision change the energy per token,
and how this energy splits between prefill and decode. At the layer level, we
look at how the energy of each phase is spread across layer roles;
that minimize the energy per token.

\textbf{Architecture level.}
We start from two reference configurations and vary one factor at a time
around each: a \emph{small} configuration (batch size 1, 16-token prompt,
16 generated tokens, bf16) and a \emph{large} one (batch size 64, 256-token
prompt, 64 generated tokens, bf16). Figure~\ref{fig:oat} shows the energy
per token of each model for the four factors.
\textbf{Batch size} has by far the largest effect. From batch size 1 to 128, the
energy per token drops by about two orders of magnitude, e.g. from about 9
to 0.15~J/token for Qwen3-8B around the small configuration. Larger batches
process more requests in parallel and share the cost of each decode step
across them. The trend holds around both configurations, until the GPU
memory is full.
The \textbf{prompt length} has a different 
impact around the two configurations. Around
the small one, the energy per token stays almost flat up to
256--1024 prompt tokens, then rises sharply: with 8192 tokens,
Qwen3-8B reaches about 40~J/token, 4.4$\times$ its value with 16 tokens.
A longer prompt makes the prefill more expensive
and it also enlarges the KV cache read at every decode step. Around the
large configuration there is almost no plateau, and the energy per token
increases from the shortest prompts on.
The \textbf{number of generated tokens} distinguishes the two configurations most
clearly. Around the small configuration, the energy per token is nearly
constant (5.9 to 6.8~J/token for Qwen3-0.6B from 16 to 128 tokens): almost
all the energy is spent in decode, and each new token costs about the same.
Around the large configuration, the prefill accounts for a large share of
the energy, and this fixed cost is divided by more tokens as generation goes
on. The energy per token decreases accordingly, e.g. from 0.64 to
0.29~J/token for Qwen3-4B between 16 and 64 tokens.

Our setup covers three \textbf{precisions}: fp32, fp16 and
bf16. fp16 and bf16 give almost the same energy per token, as both store an
element in two bytes. fp32 costs more for most models around the small
configuration, up to about +50\% for Qwen3-8B and Llama-3.2-3B, while the
smallest models are barely affected.
Around the large configuration, the effect is much stronger even for small models: fp32 multiplies the energy per token
by 1.9 to 2.8 for the three models that fit in memory.

Wattribution measures the energy of the prefill and decode phases within each run.
The split %
confirms these trends
(see Figure~\ref{fig:phase_share}), where each bar averages all the runs that
share its two settings. A longer prompt increases the prefill share,
and more so at larger batch sizes: with a 256-token prompt, the prefill
accounts on average for 5\% of the energy at batch size 1 and 47\% at
batch size 128. 

Generating more tokens has the opposite effect, since the fixed prefill cost is spread over more decode steps. 
With an 8192-token prompt, the prefill
represents on average 66\% of the energy of a run that generates 16 tokens, but only
22\% when it generates 128. 
These results support splitting the energy prediction between prefill and decode, as
done in our estimation model (Section~\ref{sec:estimation}).
They also point to where the energy per token is minimal: a batch
size as large as memory allows and a controlled prompt length keep the cost
of a token low and nearly constant.

\setcounter{figure}{4}
\begin{figure}[t]
    \centering
    \includegraphics[width=\linewidth]{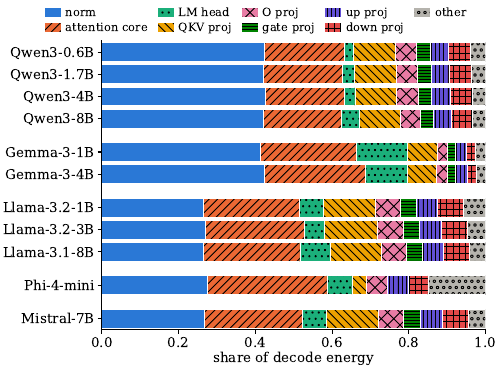}
    \caption{Mean share of each layer in the decode energy.}
    \label{fig:role_share}
\end{figure}

\begin{table*}[t]
\centering
\caption{
  Prediction accuracy on 1{,}548 runs under cross-configuration (CC), leave-one-model-out (LOMO) and leave-one-familly-out (LOFO) validation, compared to baseline models.
  Ranking metrics on the total energy only, restricted to the
  159 of 200 configurations run by at least five models. 
  Pair accuracy is computed per configuration and ignores pairs of models whose measured energies differ by less than 5\%.
}
\label{tab:accuracy}
\small
\begin{tabular*}{\textwidth}{@{\extracolsep{\fill}}lrrrrrrrr}
\toprule
Method & \shortstack{Median\\signed (\%)} & \shortstack{Std dev of\\signed (\%)} & \shortstack{MedAPE\\(\%)} & \shortstack{MAPE\\(\%)} & \shortstack{$\rho_{\mathrm{all}}$\\all runs} & \shortstack{$\rho_{\mathrm{model}}$\\median/model} & \shortstack{$\rho_{\mathrm{config}}$\\median/config} & \shortstack{Pair acc.\\(\%)} \\
\midrule
\multicolumn{9}{l}{\emph{Cross-configuration (5-fold grouped CV)}} \\
B-flops & +26.8 & 161.6 & 49.0 & 107.9 & 0.557 & 0.575 & 0.882 & 90.0 \\
B-params & $-$3.0 & 35.3 & 18.9 & 25.4 & 0.928 & 0.969 & 0.905 & 90.0 \\
B-roof & +9.0 & 46.6 & 28.2 & 36.4 & 0.909 & 0.965 & 0.873 & 89.5 \\
Wattribution & \textbf{$-$0.6} & \textbf{9.3} & \textbf{4.8} & \textbf{6.7} & \textbf{0.993} & \textbf{0.992} & \textbf{0.982} & \textbf{100.0} \\
\midrule
\multicolumn{9}{l}{\emph{Leave-one-model-out}} \\
B-flops & +29.0 & 169.9 & 50.1 & 111.1 & 0.507 & 0.605 & 0.809 & 84.3 \\
B-params & $-$4.3 & 44.2 & 22.5 & 31.4 & 0.887 & 0.970 & 0.809 & 83.3 \\
B-roof & +11.0 & 49.9 & 32.2 & 39.7 & 0.897 & 0.968 & 0.857 & 88.9 \\
Wattribution & \textbf{+0.5} & \textbf{20.8} & \textbf{13.1} & \textbf{16.0} & \textbf{0.969} & \textbf{0.983} & \textbf{0.893} & \textbf{90.0} \\
\midrule
\multicolumn{9}{l}{\emph{Leave-one-family-out}} \\
B-flops & +22.0 & 167.1 & 51.4 & 108.0 & 0.514 & 0.606 & 0.714 & 77.8 \\
B-params & $-$7.7 & 47.2 & 26.1 & 34.6 & 0.870 & \textbf{0.972} & 0.736 & 79.6 \\
B-roof & +13.0 & 51.9 & 34.2 & 41.9 & 0.886 & 0.968 & 0.809 & 83.3 \\
Wattribution & \textbf{+4.9} & \textbf{42.1} & \textbf{16.9} & \textbf{25.0} & \textbf{0.939} & 0.961 & \textbf{0.845} & \textbf{86.3} \\
\bottomrule
\end{tabular*}
\end{table*}
\textbf{Layer-wise attribution} also shows how the energy of each phase is spread across the layers of the model. Figure~\ref{fig:role_share} shows the share of the decode energy attributed to each role. Normalization layers receive the largest share, about 37\% on average, followed by the attention core with about 24\%, while the linear projections receive a smaller share. The share of normalization layers is larger in Qwen3 and Gemma~3, which contain more of them per block.

\noindent\colorbox{gray!10}{\parbox{\dimexpr\columnwidth-2\fboxsep\relax}{%
\textbf{Answering RQ2: What are the main trends and mechanisms behind the energy consumption of a language model?}
Batch size is the main lever on the energy per token: grouping requests shares the cost of each decode step, and the energy per token drops by up to 60 times from batch size 1 to 128.
The other settings act through the balance between prefill and decode. Long prompts make the prefill more expensive and enlarge the KV cache read at every decode step, so that the prefill accounts for up to two thirds of the energy of a run, whereas generating more tokens spreads this fixed cost over more decode steps.
Precision matters mainly through the bytes per element: fp16 and bf16 cost the same, while fp32 raises the energy per token by up to 2.8 times.
In practice, we therefore recommend batching requests as much as memory allows, keeping prompts short, and running in half precision.
}}

\subsection{Prediction Accuracy}
\label{sec:prediction}

Finally, we use the collected data to train the energy estimation model described in Section~\ref{sec:estimation}. One estimation function is trained per layer type (\texttt{lm\_head}, \texttt{norm}, \texttt{v\_proj}, etc.) and per phase (prefill and decode). 
As we aim for a model that is lightweight, easy to train and interpretable, we compared several model families (linear, log-linear, small multilayer perceptrons and tree-based models) and tuned the
hyperparameters of each estimation function separately.
Consistent with \cite{argerich_wattgpu_2026}, tree-based models give the best accuracy: we retain the best model for each layer type and phase, and most of the selected models are random forests or XGBoost (more than $9$ of the $14$ estimation functions). 
We evaluate the resulting model under three validation schemes.
Cross-Configuration (CC) validation holds out configurations: 5-fold grouped cross-validation over the 200 inference configurations: each fold holds out 40 configurations for all models, so every model is seen in training but the test combinations are not (batch, prompt, generation, precision) in order and tests whether the model generalizes to unseen inference settings. 
Leave-one-model-out (LOMO) validation holds out one language model at a time and tests whether
it generalizes to unseen architectures \cite{argerich_wattgpu_2026}.
Leave-one-family-out (LOFO) validation holds out one family at a time and tests whether
it generalizes to unseen architectures \cite{argerich_wattgpu_2026}.
For all, we report two complementary metrics: the relative error of the predicted energy and the ability of the predictions to rank runs by energy (see Table~\ref{tab:accuracy}).

\textbf{Accuracy Evaluation.}
The first metric assesses the absolute performance of our model: can we accurately predict the energy of a model before executing it, and does this prediction generalize to language models never seen during training?
In cross-configuration validation, \textbf{the MedAPE is 4.8\%}, with a median signed error of $-0.6$\%. When an entire model is removed from the training set, i.e. \textbf{in LOMO validation, the MedAPE rises to 13.1\%}, and to \textbf{16.9\% when a whole family is removed (LOFO)}, while the median signed error stays close to zero ($+0.5$\% and $+4.9$\%).
In every setting, Wattribution is far more accurate than the baselines: the best of them, based on the number of parameters, reaches a MedAPE of 18.9\%, 22.5\% and 26.1\%, and the FLOPs and roofline baselines exceed 28\%.
Behind these figures lie two estimation models, one for the prefill and one for the decode. The decode estimation performs well across settings, while the prefill is harder to predict (MedAPE of 28.7\% against 10.2\% for the decode in LOMO validation), for several reasons.
(i) The prefill happens only once per inference, so far fewer samples are available to train its estimation model.
(ii) As shown in Section~\ref{sec:energy}, the prefill energy varies strongly across configurations. The same model must therefore generalize across very different settings, whereas the decode behaves more consistently.

\textbf{Ranking performance.}
We evaluate the agreement between predicted and measured rankings with the rank correlation coefficient $\rho$ of \cite{spearman1904association} and with the pair accuracy, the share of pairs of runs that the model orders correctly.
We compute them at three aggregation levels, each answering one question:
(i) over the entire pool, does the model order all held-out runs by energy?
(ii) within a model, does it order its configurations (batch size, prompt length, precision, etc.) correctly?
(iii) within a configuration, does it pick the cheapest model?
The model ranks runs well. In cross-configuration validation, $\rho$ is at least $0.98$ at every level.
As expected, the ranking degrades on unseen models, but only slightly: over the entire pool, $\rho$ goes from $0.993$ to $0.969$ in LOMO validation, and to $0.939$ in LOFO validation.
At the two finer levels, the model orders the configurations of a given model more easily than the models of a given configuration ($\rho$ of $0.983$ against $0.893$ in LOMO validation).
The pair accuracy remains at 90\% in LOMO validation and 86\% in LOFO validation: the model still picks the cheaper of two models in most cases.
Our model can therefore be used not only to predict the energy consumption of a language model, but also to rank models and choose the most efficient one for a given configuration.

\noindent\colorbox{gray!10}{\parbox{\dimexpr\columnwidth-2\fboxsep\relax}{%
\textbf{Answering RQ3: How can we predict, without running it, how much energy a given language model will consume?}
Wattribution estimation model meets its two design goals.
It is lightweight: its features are computed analytically from the model configuration.
It is transferable: the MedAPE is 4.8\% on unseen configurations, 13.1\% on unseen language models and 16.9\% on unseen model families.
Finally, the predictions preserve the energy ranking of runs ($\rho = 0.969$ in LOMO validation), and the model picks the cheaper of two language models in 90\% of the pairs, which makes it usable to choose the most efficient one before deployment.
}}

\section{Discussion}
\label{sec:discussion}

Collecting energy data on processes shorter than the resolution of the sensor raises a question that precedes any estimation: which power reading to use, and how to attribute it. 
Our characterization of the NVML readings and our Wattribution protocol answer those questions.
We combine the finest available reading with an attribution performed within the actual execution of the model, without any correction factor. 
This principle is not specific to layers: the attribution only relies on timestamped windows and on averaging over repetitions. It applies to any sub-process that can be delimited during execution. 
Building the dataset required a measurement campaign of 3.4~kWh, but this cost is paid only once: afterwards, the energy of a new configuration or of an unseen model is estimated instantly, enabling fast decisions at design time. 
Wattribution balances precision, overhead and portability. It relies only on the power sensor built into recent NVIDIA GPUs, so that it runs wherever the model runs, including shared clusters and cloud instances where no external instrument can be installed. 

We plan to extend this procedure in three directions. First, our focus was on the GPU, which accounts for most of the energy of LLM inference \cite{argerich_measuring_2024}.
The CPU, nevertheless, contributes a non-negligible share, and the same protocol could collect its energy in parallel, or when inference runs on the CPU alone. 
Second, the measurement campaign runs on a single GPU model. Since the sample buffer offers a similar resolution on the nine GPUs we characterized, the protocol can be applied to them directly, to study how the energy of language models transfers across hardware. 
Third, our measurements rely on Hugging Face Transformers, and we plan to adapt Wattribution to inference engines such as vLLM or TensorRT-LLM.

\section{Conclusion}
\label{sec:conclusion}

We introduced Wattribution, a protocol and tool that attributes the energy of a language model to each of its layers within the actual execution, building on the first characterization, to our knowledge, of the high-resolution power samples exposed by NVIDIA GPUs.
The collected data show that the energy per token is lowest with a batch size as large as memory allows, short prompts and half precision.
Leveraging these measurements, our analytical, layer-wise estimator predicts the energy consumption of language models with a median absolute percentage error of 4.8\% on unseen configurations, 13.1\% on unseen language models, and 16.9\% on model families never seen during training, without any correction factor.
By making energy as accessible as accuracy and latency at design time, this work empowers developers and organizations to integrate energy efficiency into the choice and deployment of AI models.

\bibliography{example_paper}
\bibliographystyle{mlsys2025}

\section*{AI Usage Disclosure}

The research questions, methodology, experimental design and analyses were conceived by the authors. No generative AI tool was used for ideation. 
Cursor was used to support the development of the measurement and analysis code, for code review, implementation of tests and debugging. All code was reviewed by the authors, and all experiments and reported results were run and checked by them. 
Cursor was also used for a complementary literature review, for drafting assistance and for language editing. After using these tools, the authors reviewed and edited the content as needed and take full responsibility for the publication's content.

\appendix
\newpage

\section{The aggregation gap and the correction factor}
\label{app:aggregation-gap}
 
On GPU the sum of per-layer energies does not recover the measured energy of
the architecture. Both WattLayer and IrEne close the gap at the aggregation
step, with a multiplicative factor.
 
WattLayer~\cite{sardi2026wattlayer} reports a systematic underestimate of
approximately 25\%, corrected by a single constant,
\begin{equation}
\label{eq:alpha}
E_{\mathcal{A}} \;\approx\; \alpha \sum_{i=1}^{N} E_{L_i} ,
 \alpha = 1.25 ,
\end{equation}
calibrated once and found stable across hardware platforms and tasks.
 
IrEne~\cite{cao_etal_2021_irene} replaces the constant with a learned per-child
multiplier. Writing $P_e(n)$ for the predicted energy of node $n$ and
$\mathrm{feat}(c)$ for the features of child $c$,
\begin{equation}
    P_e(n) \;=\; \sum_{c \,\in\, \mathrm{children}(n)} \alpha(c) \, P_e(c) , \\
\end{equation}
\begin{equation}
    \alpha(c) \;=\; 1 + \frac{\tanh\!\bigl(W \cdot \mathrm{feat}(c) + b\bigr)}{\tau} ,
\end{equation}
with $W$ and $b$ learned and $\tau$ fixed, so each weight stays within
$1 \pm 1/\tau$ of unity.
 
Both act on the sum and leave the per-layer quantity unexamined. Among IrEne's
resource features are the GPU core and memory clocks: the clock already appears
in the literature as a predictor of the correction, not as its cause.
 
\subsection{Protocol}
\label{app:depth-sweep}

We design the following experiment. Its goal is to determine how much energy a
single fully connected layer consumes, and what the architecture containing it
contributes to that figure.

The architecture is a stack of $k$ identical fully connected layers of hidden
dimension 512, with no other operation present. Depth is swept over
$k \in [1, 11]$; each configuration is executed 1028 times with GPU power
sampled at 100\,Hz, and the per-layer energy is taken as the increment from
adding one layer to the stack. The sweep is then repeated in the opposite
direction over $k \in [1, 11]$, with every other parameter unchanged. Since all
layers are identical, variation across $k$ cannot be a property of the layer.

\subsection{Result}
 
\begin{figure}[t]
\begin{center}
\centerline{\includegraphics[width=\columnwidth]{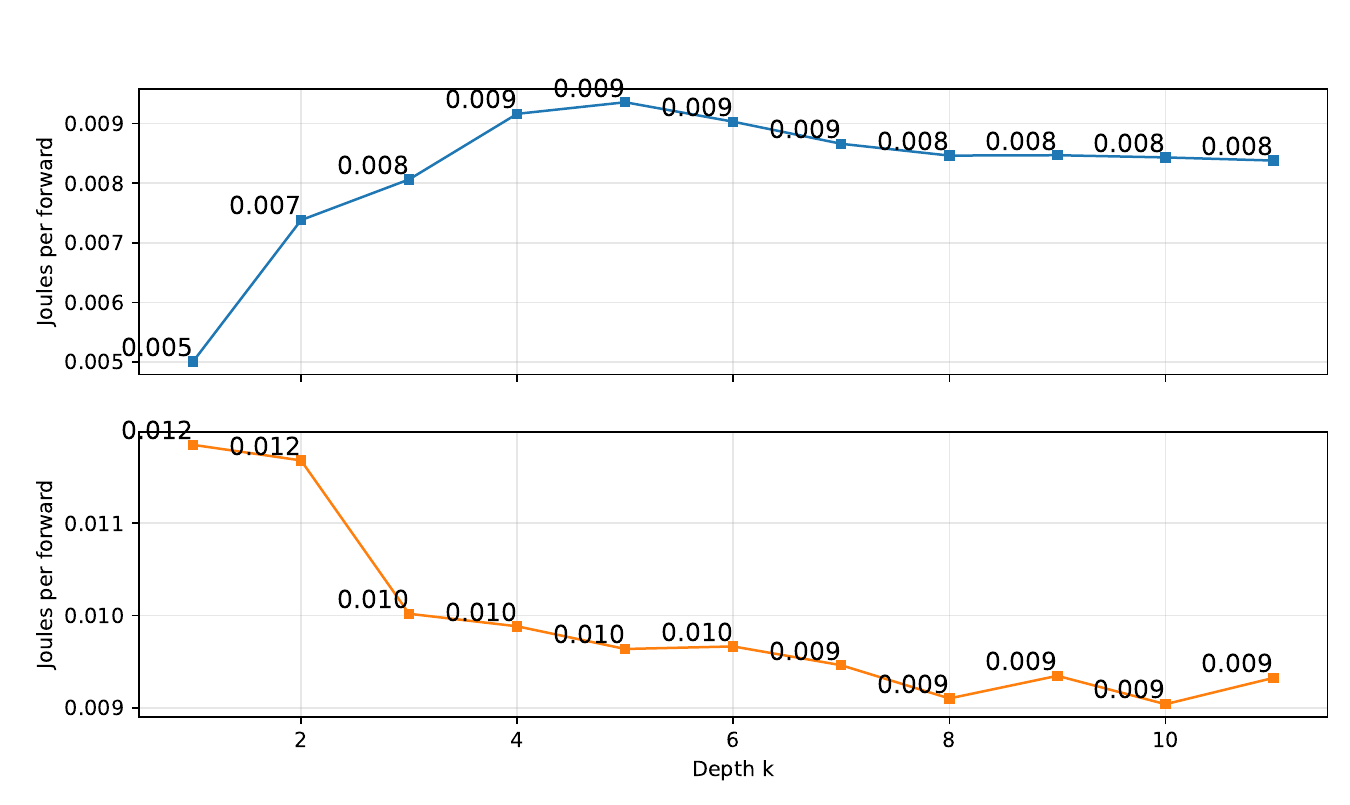}}
\caption{Per-layer energy against stack depth $k$, ascending (top) and
descending (bottom). The layers are identical; only the measurement order
differs. The two runs disagree by a factor of 2.4 at $k = 1$ and converge as
the stack grows.}
\label{fig:energy-depth-appendix}
\end{center}
\vskip -0.2in
\end{figure}

Figure~\ref{fig:energy-depth-appendix} shows the two sweeps. The ascending run
rises from 0.005\,J at $k = 1$ to settle near 0.008\,J; the descending run
starts at 0.012\,J and falls toward 0.009\,J. Two conclusions follow.
 
\textbf{Isolated execution does not reach the operating frequency of the
device.} A stack of one layer completes before the clock governor has ramped,
so the layer executes at a low frequency and draws correspondingly little
power. Run in reverse, the device enters the same configuration already at its
sustained frequency and holds it, and the identical layer then draws twice as
much. The frequency at which a layer executes is set by the work surrounding it
and by the order of the measurement campaign, not by the layer. An isolated
measurement therefore does not return the energy of the layer: it returns the
energy of that layer in a regime that does not occur during inference, where
the layer runs inside a complete architecture at sustained clock.

Equation~\ref{eq:alpha} follows. Isolated execution places every layer at
$k = 1$, understates each of them, and understates the small ones most; a
constant above unity is what is needed to restore the total. On this reading
$\alpha = 1.25$ is the aggregate footprint of a measurement artifact rather
than a property of the architectures, and the appropriate response is to change
the measurement rather than to calibrate the correction.

\begin{table*}[t]
\centering
\footnotesize
\caption{Operations of one Transformer block, in execution order, for a
generation step processing $q$ tokens per sequence. Values are obtained by
applying the definitions of Section~\ref{app:features_def} to each operation.
Operations that are not layers (no check mark) have no row in the feature set.
The table highlights the two kinds of model features: the computation
performed by each operation (\emph{compute}) and the bytes it moves
(\emph{memory}), split into weights, inputs read, and outputs written.
$^\dagger$Values are given per projection.}
\label{tab:block_ops}
\setlength{\tabcolsep}{4pt}
\begin{tabular*}{\textwidth}{@{\extracolsep{\fill}}llccccc}
\toprule
  & & & {Compute} & \multicolumn{3}{c}{Memory (bytes)} \\
\cmidrule(lr){4-4} \cmidrule(l){5-7}
Operation & Type & Layer & MACs & Weights & Inputs & Outputs \\
\midrule
\multicolumn{7}{@{}l}{\emph{Attention sub-block}} \\
\quad\texttt{input\_layernorm}                    & element-wise & \checkmark & $\approx 0$         & $d\,b_e$             & $Bq\,d\,b_e$       & $Bq\,d\,b_e$ \\
\quad\texttt{q\_proj}                             & linear       & \checkmark & $d\,h d_h\,Bq$      & $d\,h d_h\,b_e$      & $Bq\,d\,b_e$       & $Bq\,h d_h\,b_e$ \\
\quad\texttt{k\_proj}, \texttt{v\_proj}$^\dagger$ & linear       & \checkmark & $d\,h_{kv} d_h\,Bq$ & $d\,h_{kv} d_h\,b_e$ & $Bq\,d\,b_e$       & $Bq\,h_{kv} d_h\,b_e$ \\
\quad\texttt{o\_proj}                             & linear       & \checkmark & $h d_h\,d\,Bq$      & $h d_h\,d\,b_e$      & $Bq\,h d_h\,b_e$   & $Bq\,d\,b_e$ \\
\quad residual addition                           & element-wise &            & $\approx 0$         & --                   & $2\,Bq\,d\,b_e$    & $Bq\,d\,b_e$ \\
\addlinespace
\multicolumn{7}{@{}l}{\emph{MLP sub-block}} \\
\quad\texttt{post\_attention\_layernorm}          & element-wise & \checkmark & $\approx 0$         & $d\,b_e$             & $Bq\,d\,b_e$       & $Bq\,d\,b_e$ \\
\quad\texttt{gate\_proj}, \texttt{up\_proj}$^\dagger$ & linear   & \checkmark & $d\,d_{ff}\,Bq$     & $d\,d_{ff}\,b_e$     & $Bq\,d\,b_e$       & $Bq\,d_{ff}\,b_e$ \\
\quad\texttt{act\_fn} (SiLU)                      & element-wise & \checkmark & $\approx 0$         & --                   & $Bq\,d_{ff}\,b_e$  & $Bq\,d_{ff}\,b_e$ \\
\quad gating product                              & element-wise &            & $\approx 0$         & --                   & $2\,Bq\,d_{ff}\,b_e$ & $Bq\,d_{ff}\,b_e$ \\
\quad\texttt{down\_proj}                          & linear       & \checkmark & $d_{ff}\,d\,Bq$     & $d_{ff}\,d\,b_e$     & $Bq\,d_{ff}\,b_e$  & $Bq\,d\,b_e$ \\
\quad residual addition                           & element-wise &            & $\approx 0$         & --                   & $2\,Bq\,d\,b_e$    & $Bq\,d\,b_e$ \\
\bottomrule
\end{tabular*}
\end{table*}

\section{Model Features}
\label{app:features}

This appendix details the model features of Section~\ref{sec:estimation}. Section~\ref{app:features_structure} introduces the structure of a language model and the notation. Section~\ref{app:features_block} describes the computation and memory traffic of each operation of a Transformer block, and why prefill and decode follow different regimes. Section~\ref{app:features_pipeline} describes how the features are extracted, and Section~\ref{app:features_def} defines each of them. All features are computed analytically from the configuration of the model: no weights are downloaded and no computation runs on the GPU, so the features of a model can be obtained before it is ever executed.

\subsection{Structure of a language model and notation}
\label{app:features_structure}

A language model architecture $\mathcal{A}$ maps each input token to a vector of size $d$, called the hidden size, through an embedding layer. This vector then goes through a stack of $N_b$ identical \emph{Transformer blocks}, and a final linear layer, the language-modeling head, projects the result onto a vocabulary of $n_v$ tokens to predict the next one. Each Transformer block contains two sub-layers, each preceded by a normalization and followed by a residual addition. The first one is the \emph{attention} sub-layer, in which each token gathers information from the previous tokens. It uses $h$ query heads of dimension $d_h$ and $h_{kv} \leq h$ key--value heads, $h_{kv} < h$ corresponding to grouped-query attention. The second one is the \emph{MLP} sub-layer, in which each token is transformed independently: its vector is projected from size $d$ to a larger size $d_{ff}$, then back to size $d$. In the terms of Section~\ref{sec:wattribution}, each block contains several layers $L_i$: its linear projections, its normalization layers and its activation function.

We consider the inference of a batch of $B$ sequences, each with a prompt of $S$ tokens, from which the model generates $T$ new tokens. The numerical precision sets the number of bytes per element $b_e$, for instance $b_e = 4$ in fp32 and $b_e = 2$ in fp16 and bf16. Generation proceeds in $T$ steps $s = 0, \dots, T-1$. Each step is characterized by the number of tokens $q$ it processes per sequence and by the length $C$ of the context the attention attends to.
\begin{itemize}
    \item \textbf{Prefill} ($s = 0$): the model processes the whole prompt at once and produces the first token, so $q = S$ and $C = S$.
    \item \textbf{Decode} ($s = 1, \dots, T-1$): the model processes one new token per step, so $q = 1$ and $C = S + s$.
\end{itemize}
The first generated token is attributed to prefill, so there are $T - 1$ decode steps.

\subsection{Computation and memory in a Transformer block}
\label{app:features_block}

Table~\ref{tab:block_ops} lists the operations of one Transformer block in execution order, with their computation, counted in multiply--accumulate operations (MACs), and their memory traffic.

Three kinds of operations appear. The linear layers carry most of the parameters: their computation scales with the number of tokens $Bq$, and they read their full weight matrix at every step. The attention core, which computes $QK^\top$, the softmax and the product with $V$, has no weights. Instead, it reads the keys and values of the whole context, stored in the KV cache, so its cost grows with $C$. 
Finally, element-wise operations, such as normalizations, activations and residual additions, perform little computation and only move activations.

\textbf{Prefill and decode regimes.}
The balance between computation and memory traffic is measured by the operational intensity, the number of MACs performed per byte read from memory. A layer with a high intensity is limited by the computing power of the GPU (compute-bound), whereas a layer with a low intensity spends most of its time waiting for memory (memory-bound). For a linear layer, the weights dominate the memory traffic, and its intensity is about $Bq/b_e$. During prefill, $q = S$: each weight read from memory is reused for the $BS$ tokens of the batch, and the layer is compute-bound. During decode, $q = 1$: the full weight matrix is read for only $B$ tokens, and the same layer becomes memory-bound. The attention core follows a third behavior. During decode, its intensity is about $h/(h_{kv}\,b_e)$, of the order of one MAC per byte, whatever the batch size. It is therefore memory-bound at every decode step, and its traffic grows with each generated token, since $C = S + s$. The same layer thus operates in two physical regimes depending on the phase, which is why the features are computed separately for prefill and decode.

\subsection{Extraction pipeline}
\label{app:features_pipeline}

The extraction proceeds in three stages.

\textbf{Structural trace.}
We build the architecture from its Hugging Face configuration, on the \texttt{meta} device or on CPU, so that no weights are loaded. We then run a single forward pass on a short dummy input and attach a hook to every layer, that is, every module without children (similar to \cite{sardi2026wattlayer}). 
The hooks record the order in which the layers fire, their names and the shapes of their inputs and outputs. From these shapes we obtain, for each layer, its input dimension $d_{\mathrm{in}}$, its output dimension $d_{\mathrm{out}}$ and its number of parameters. Tracing the actual forward pass, rather than listing the modules declared by the model, ensures that each row corresponds to a layer that really executes, in the order in which it executes. Each layer is also assigned a role from its name, such as \texttt{q\_proj}, \texttt{o\_proj}, \texttt{up\_proj} or \texttt{norm}. The attention dimensions are read from the configuration: the number of attention heads $h$, the number of key--value heads $h_{kv}$, which covers grouped-query and multi-query attention, and the head dimension $d_h$.

\textbf{Analytical rescaling.}
The trace uses a fixed dummy input, whose shape only serves to recover the dimensions of the layers. The features are then computed for the configuration of the experiment, defined by the batch size $B$, the prompt length $S$, the number of generated tokens $T$ and the precision, through the closed-form expressions of Section~\ref{app:features_def}. All features are first computed per layer and per step. The prefill features of a layer are those of step $0$, and its decode features are the sum over the decode steps,
\begin{equation}
x_{\ell}^{\text{decode}} = \sum_{s=1}^{T-1} x_{\ell}(s),
\end{equation}
which is consistent with the aggregation of the measured energies (Equation~\ref{eq:aggregation}). The weight bytes are the only exception: they do not depend on the step and are not summed.

\textbf{Correctness check.}
For every matrix-multiplication layer, we verify that the analytical MACs and activation counts, evaluated at the dimensions of the dummy input, match the values recorded by the hooks during the trace.

\subsection{Feature definitions}
\label{app:features_def}

Table~\ref{tab:features} summarizes the model features. For a layer at a given step, we write $n = Bq$ for the number of tokens processed and $b_e$ for the number of bytes per element.

\begin{table}[t]
\centering
\footnotesize
\caption{Model features computed for each layer and each generation step. $d_{\mathrm{in}}$ and $d_{\mathrm{out}}$ are the input and output dimensions of the layer; the other symbols follow Section~\ref{app:features_structure}.}
\label{tab:features}
\setlength{\tabcolsep}{4pt}
    \begin{tabular*}{\columnwidth}{@{\extracolsep{\fill}}ll}
\toprule
\textbf{Feature} & \textbf{Definition} \\
\midrule
MACs & $d_{\mathrm{in}}\, d_{\mathrm{out}}\, B q$ \\
Weight bytes & $\#\text{params} \times b_e$ \\
Input bytes & $B q\, d_{\mathrm{in}}\, b_e$ \\
Output bytes & $B q\, d_{\mathrm{out}}\, b_e$ \\
DRAM bytes & input + output + weight bytes \\
Operational intensity & MACs / DRAM bytes \\
Attention core MACs & $2 B h\, q\, C\, d_h$ \\
KV cache bytes & $2 B h_{kv}\, C\, d_h\, b_e$ \\
Bytes per element & $b_e$ \\
\bottomrule
\end{tabular*}
\end{table}

\textbf{MACs.}
The number of multiply--accumulate operations performed by a dense projection is
\begin{equation}
\mathrm{MACs} = d_{\mathrm{in}}\, d_{\mathrm{out}}\, n.
\end{equation}
Layers that perform no matrix multiplication, such as normalization layers, activation functions, embeddings and rotary position encodings, are assigned zero MACs. The matrix multiplications of the attention mechanism itself are counted separately (see the attention core MACs below).

\textbf{Input and output bytes.}
For a matrix-multiplication layer, the bytes read as input and written as output are
\begin{equation}
\mathrm{Bytes}_{\mathrm{in}} = n\, d_{\mathrm{in}}\, b_e, \qquad \mathrm{Bytes}_{\mathrm{out}} = n\, d_{\mathrm{out}}\, b_e.
\end{equation}
For the other layers, the hidden width $w$ is recovered from the activations recorded during the trace, and the input and output bytes are both set to $n\, w\, b_e$.

\textbf{Weight bytes.}
The bytes occupied by the parameters of the layer are
\begin{equation}
\mathrm{Bytes}_{\mathrm{w}} = \#\mathrm{params} \times b_e.
\end{equation}
We assume that the weights are read once at every step.

\textbf{DRAM bytes.}
The memory traffic of a layer is approximated as the sum of the three previous quantities,
\begin{equation}
\mathrm{Bytes}_{\mathrm{DRAM}} = \mathrm{Bytes}_{\mathrm{in}} + \mathrm{Bytes}_{\mathrm{out}} + \mathrm{Bytes}_{\mathrm{w}}.
\end{equation}

\textbf{Operational intensity.}
Following the roofline model \cite{williams_roofline_2009}, the operational intensity of a layer is the ratio between its computation and its memory traffic,
\begin{equation}
I = \frac{\mathrm{MACs}}{\mathrm{Bytes}_{\mathrm{DRAM}}}.
\end{equation}
A low intensity indicates a memory-bound layer, a high intensity a compute-bound one. It is undefined for layers with no memory traffic.

\textbf{Bytes per element.}
The number of bytes per element $b_e$ is also used directly as a feature, as it encodes the precision of the run.

\section{Power Sensor Characterization}
\label{app:sensor}

\begin{figure*}[h]
    \centering
    \includegraphics[width=.9\linewidth]{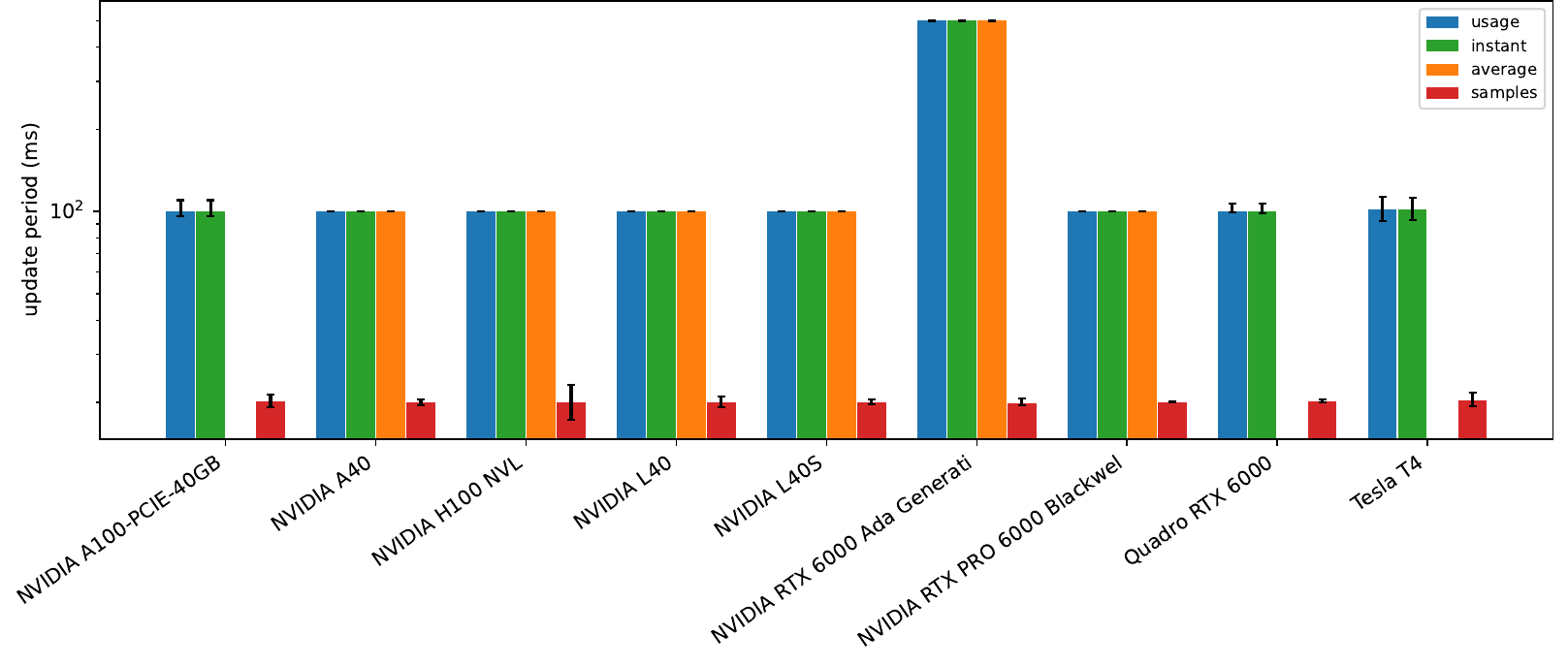}
    \caption{Update period of each NVML reading at 10~ms polling on the nine GPUs (median and 10th--90th percentiles). \texttt{samples} is the fastest reading on every unit; \texttt{usage} is refreshed every 100~ms on the cluster GPUs and every 500~ms on the reference workstation.}
    \label{fig:app_update_period}
\end{figure*}

\begin{figure*}[t]
\centering
\includegraphics[width=.9\textwidth]{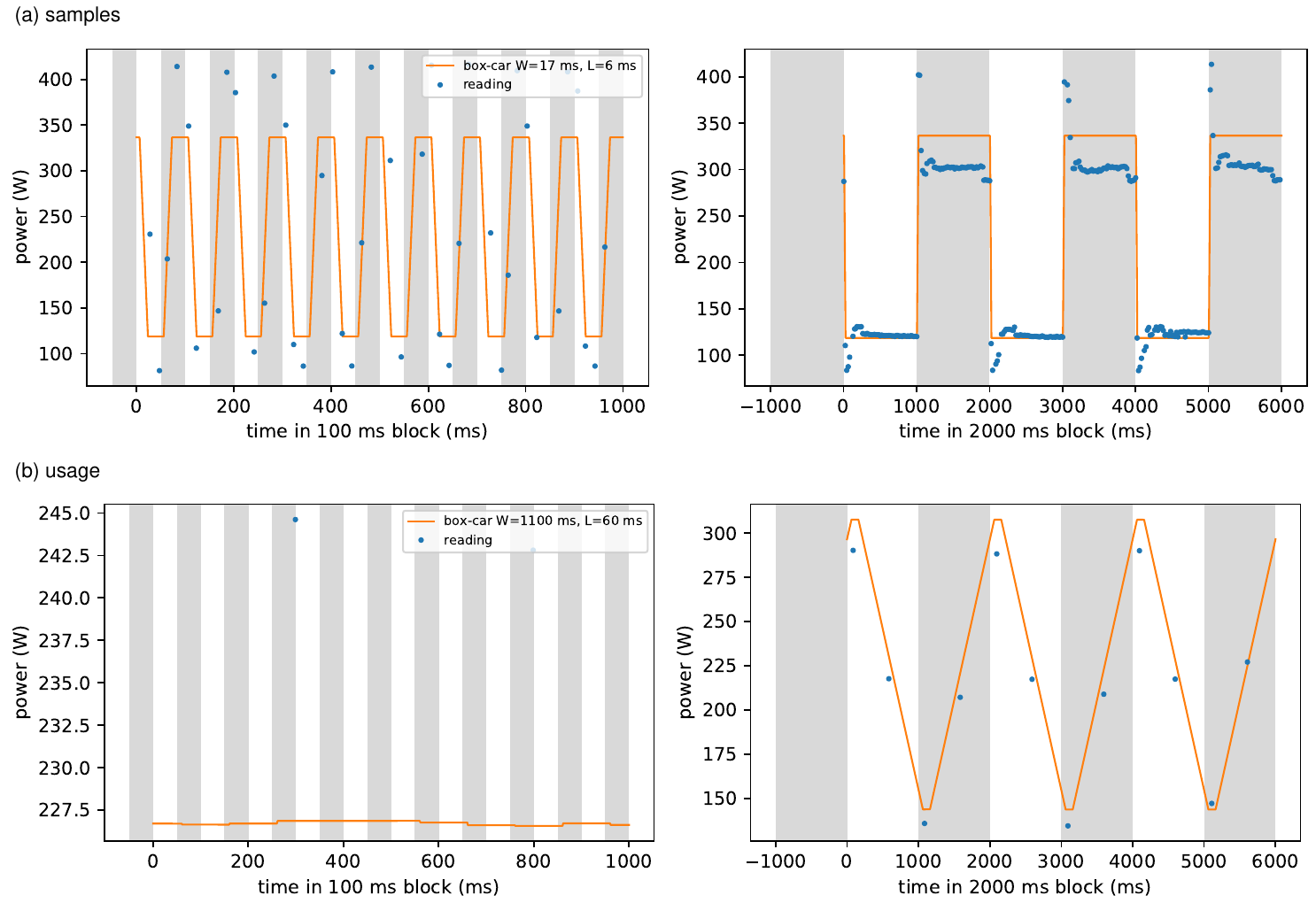}
\caption{\texttt{samples} (a) and \texttt{usage} (b) under the square-wave load on the reference workstation (grey: load on). Left: 100~ms period. Right: 2~s period. Dots are readings; lines are the predictions of Equation~\ref{eq:reading} with the fitted window and lag. \texttt{samples} produces a reading about every 20~ms, averaged over $W = 17$~ms with a lag $L = 6$~ms ($R^2 = 0.87$), and follows each 100~ms burst. \texttt{usage} produces a reading every 500~ms, averaged over about 1.1~s ($W = 1100$~ms, $L = 60$~ms, $R^2 = 0.86$): the 100~ms square wave is invisible, and the 2~s square wave is rendered as a triangle, because each reading mixes on and off time.}
\label{fig:square_wave}
\end{figure*}

This appendix details the characterization of the NVML power readings summarized in Section~\ref{sec:sensor}.

\subsection{Probe protocol}
\label{app:sensor_protocol}

A probe is a short measurement of the sensor alone, without any language model. A script polls the five NVML readings in a loop at a fixed interval (2, 5 or 10~ms) and records every returned value together with the host monotonic clock, read just before and just after the call. Meanwhile, the GPU is in one of three known states: \emph{idle}, with nothing running; \emph{load}, a continuous fp16 matrix multiplication; and \emph{square wave}, the same multiplication switched on and off with a 50\% duty cycle, in blocks with periods from 10~ms to 4~s, each block starting at a random offset. Every on/off transition is timestamped on the same host clock as the readings, so the true load signal is known to the millisecond, and any discrepancy between the readings and the load can be attributed to the sensor.

On the reference workstation, we collected seven probe runs: idle and load at 10~ms and 2~ms polling, idle and load at 10~ms on a second day, and one square-wave run at 5~ms polling. 
On the eight cluster GPUs, each campaign comprises idle and load at 10~ms and 2~ms polling and one square-wave run at 5~ms polling
In addition, we examined the sensor during real inference on the reference workstation, with two runs identical except for the power reading recorded by the sampler, \texttt{usage} or \texttt{samples}. Each run covers seven language models, from a small test model to Qwen3-8B, with batch size 32, 64 input tokens, 32 generated tokens and fp16 precision.

\subsection{Square-wave fit}
\label{app:sensor_fit}

Under the square-wave load, the true power is well approximated by a two-level signal $p(t) = a + b\,u(t)$, where $u(t) \in \{0, 1\}$ is the known on/off signal, $a$ the idle power and $b$ the additional power when the load is on. By Equation~\ref{eq:reading}, reading $k$ is then
\begin{equation}
\hat{p}_k = a + b\, x_k(W, L),
\label{eq:boxcar}
\end{equation}
with
\begin{equation}
x_k(W, L) = \frac{1}{W} \left| [t_k - L - W,\ t_k - L] \cap \mathcal{I} \right|,
\end{equation}
where $\mathcal{I}$ is the set of intervals during which the load is on, so that $x_k \in [0, 1]$ is the fraction of the window of reading $k$ during which the load was on. For a fixed pair $(W, L)$, Equation~\ref{eq:boxcar} is a simple linear regression, solved in closed form:
\begin{equation}
b = \frac{\sum_k (x_k - \bar{x})(\hat{p}_k - \bar{p})}{\sum_k (x_k - \bar{x})^2}, \qquad a = \bar{p} - b\,\bar{x}.
\end{equation}
We evaluate every pair of a grid and keep the one with the smallest sum of squared residuals. $W$ takes 32 values from 1~ms to 2\,000~ms, and $L$ ranges from 0 to 50~ms in 1~ms steps, then to 100~ms in 5~ms steps and to 500~ms in 25~ms steps. None of the selected values lies on the edge of the grid. Near the optimum for \texttt{samples}, the resolution is 2 to 3~ms for $W$ and 1~ms for $L$.

We report the quality of the fit with the coefficient of determination $R^2$, the share of the variance of the readings that the model reproduces. $R^2 = 1$ means that the model predicts every reading exactly, and $R^2 = 0$ that it does no better than the mean. $R^2$ measures how faithfully a reading follows its own fitted model, not its time resolution: a one-second average is well described by a one-second window, yet it cannot resolve anything shorter than a second. The unexplained variance mostly comes from features that a two-level model cannot represent, such as the power spike at each rising edge and the fluctuations within the on and off phases.

Table~\ref{tab:app_fits} reports the fit of each reading on the reference workstation. The fits of \texttt{samples} on the nine GPUs are given in Table~\ref{tab:sensor}.

\begin{table}[h]
\centering
\footnotesize
\setlength{\tabcolsep}{3pt}
\caption{Square-wave fit of each NVML reading on the reference workstation (driver 595.84). $n$ is the number of readings used in the fit; $W$ and $L$ in ms; $a$ and $a + b$ are the fitted idle and on-load power levels, in W.}
\label{tab:app_fits}
\begin{tabular*}{\columnwidth}{@{\extracolsep{\fill}}lrrrcc}
\toprule
\textbf{Reading} & $n$ & $W$ & $L$ & $R^2$ & $a$, $a+b$ \\
\midrule
\texttt{samples} & 13\,112 & 17   & 6  & 0.873 & 119, 337 \\
\texttt{usage}   & 593     & 1\,100 & 60 & 0.861 & 125, 326 \\
\texttt{average} & 593     & 1\,100 & 60 & 0.861 & 125, 326 \\
\texttt{instant} & 596     & 100  & 38 & 0.862 & 130, 320 \\
\bottomrule
\end{tabular*}
\end{table}

\subsection{Comparison of the readings across GPUs}
\label{app:sensor_readings}

Figure~\ref{fig:app_update_period} shows the update period of each reading on the nine GPUs. \texttt{samples} is the fastest reading on every unit, with a new value about every 20~ms. \texttt{usage}, \texttt{average} and \texttt{instant} are refreshed every 100~ms on the cluster GPUs and every 500~ms on the reference workstation. 
Figure~\ref{fig:square_wave} and Figure~\ref{fig:square_wave} show respectively \texttt{samples} and\texttt{usage} under the same square-wave load. With a 100~ms period, the bursts are invisible, and with a 2~s period, the square wave is rendered as a triangle, because each reading mixes on and off time.

\subsection{Energy counter}
\label{app:sensor_energy}

We convert the \texttt{energy} counter to power by dividing the energy accumulated between two updates by the time elapsed between them. On most data-center GPUs, the resulting power agrees with \texttt{usage}. On the Quadro RTX 6000 and on the reference workstation, it is far lower, between 2 and 25\% of the power reported by \texttt{usage} depending on the probe. Under constant load on the reference workstation, it reads about 120~W, whereas \texttt{usage} and \texttt{samples} agree at about 300~W.

\cite{dauner_evaluating_2026} report a similar underestimation on consumer GPUs when the counter is read at short intervals, but close agreement on a data-center GPU and on a workstation GPU. 
The counter brings no gain in resolution over \texttt{samples}: even where it is accurate, it is updated only every 100~ms \cite{dauner_evaluating_2026}, against about 20~ms for \texttt{samples}. We therefore do not use it.

\subsection{Polling interval}
\label{app:sensor_polling}

The polling interval must be short enough to capture every reading of the sensor, but not so short that the measurement perturbs the measured device. Following the Nyquist criterion, polling twice as fast as the sensor updates is enough to capture every reading, as recommended by \cite{bytezcom_icml_2025}. With \texttt{samples} updated about every 20~ms, we therefore poll every 10~ms.

Polling faster has a measurable cost. Comparing idle probes at 2~ms and 10~ms polling, the idle power reported by \texttt{usage} rises by up to $2.6\times$ on several cluster GPUs and by $3.1\times$ on the reference workstation. With \texttt{samples}, it stays close to $1.0\times$ on the A40, L40S and RTX PRO 6000, but rises by $2.9\times$ on the reference workstation. At 2~ms, the measurement itself keeps the GPU active and raises the power it measures. The overhead of the instrumentation during inference, including the polling thread, is quantified by experiment G1 (Section~\ref{sec:eval_wattribution}).

\subsection{Polling thread during inference}
\label{app:sensor_thread}

During the inference runs on the reference workstation, a \texttt{samples} call takes a median of 0.5~ms (99th percentile: 1.2~ms), far below the 10~ms polling interval. The polling thread paused for more than 30~ms only 14 times over about 220\,000 calls. Since \texttt{samples} is a buffer, these pauses delay readings without losing them, and only 0.2\% of about 115\,000 readings are skipped or repeated.

\end{document}